%% file: termPaper.tex
\documentclass[pdftex,12pt,a4paper]{article}

\usepackage{lipsum}

\usepackage[pdftex]{graphicx}
\usepackage{here}

\usepackage{multirow}

\usepackage{natbib}

\usepackage{amssymb}

\usepackage{amsmath}

\usepackage{geometry}
\usepackage{multirow}
\usepackage{booktabs}
\usepackage{arydshln}

\usepackage{amsthm}
\newtheoremstyle{style}   
  {0.5cm}              
  {-0.8cm}              
  {}                      
  {}                      
  {\normalfont\bfseries}  
  {{\normalfont\bfseries \thmname{#1}\thmnumber{ #2}}}
\theoremstyle{style}

\usepackage{xyling}

\usepackage{dirtytalk}
\usepackage{hyperref}

\usepackage{caption2}
\newcaptionstyle{mystyle}{%
  \normalcaptionparams
  \onelinecaptionsfalse
  \usecaptionstyle{centerlast}}

\captionstyle{mystyle}

\newcommand{\HRule}{\rule{\linewidth}{0.5mm}}

\begin{document}

\input{title.tex}

\thispagestyle{empty}
\begin{abstract}
\setlength{\parskip}{2ex plus 0.5ex minus 0.2ex}



Vision Language Models (VLMs) have demonstrated exceptional performance across various tasks. However, they have not yet been thoroughly evaluated on more complex tasks. The Persuasion Model, conceived by Aristotle, resembles a triangle shape, which highlights its inherent challenges related to personal biases. To assess the progress of VLMs on these complex tasks, we use the \texttt{ImageArg} datasets, focusing on the Logos, Ethos, and Pathos detection tasks. Our findings indicate that models from the Qwen family achieve improved F1 scores, with Qwen3 performing exceptionally well on the Logos and Pathos tasks, while Qwen2 exhibits competitive performance on the more complex Ethos detection task. We release the code to foster research in this direction \footnote{\url{https://github.com/KhondokerIslam/ArgMin.git}}. 
\end{abstract}
\newpage

\thispagestyle{empty}
\tableofcontents
\newpage

\setcounter{page}{1}		
\pagenumbering{arabic}
\newpage
\input{sections/1_introduction}

\newpage
\input{sections/2_literature_review}

\newpage
\input{sections/3_experiment}


\newpage
\input{sections/5_conclusion}

\newpage
\input{sections/6_limitations}

\newpage
\bibliographystyle{apalike}
\bibliography{references}

\end{document}

%% file: title.tex
\begin{titlepage}
\begin{center}

\includegraphics[width=0.25\textwidth]{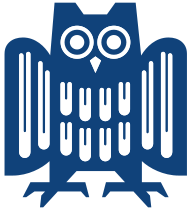}~\\[1cm]

\textsc{\LARGE Saarland  University}\\[0.4cm]
\textsc{\Large Department of Language Science \& Technology}\\[1.5cm]

\textsc{\Large Seminar:} \textbf{\Large Argumentation Mining}\\[0.5cm]

\HRule \\[1.0cm]

{ \huge \bfseries Evaluating VLMs on Multimodal Aristotelian Persuasion Tasks}\\[0.4cm]

\HRule \\[1.5cm]

\begin{minipage}{0.4\textwidth}
\begin{flushleft} \large
\emph{Author:}\\
Khondoker Ittehadul \textsc{Islam}\\
Matriculation: 7085810
\end{flushleft}
\end{minipage}
\begin{minipage}{0.4\textwidth}
\begin{flushright} \large
\emph{Supervisors:} \\
Dr. Olga \textsc{Petukhova}\\
\end{flushright}
\end{minipage}

\vfill

{\large \today}

\end{center}
\end{titlepage}

%% file: sections/1_introduction.tex
\section{Introduction}

\subsection{Persuasion Mode}

\begin{figure}[H]
\centering
\includegraphics[width=0.3\textwidth]{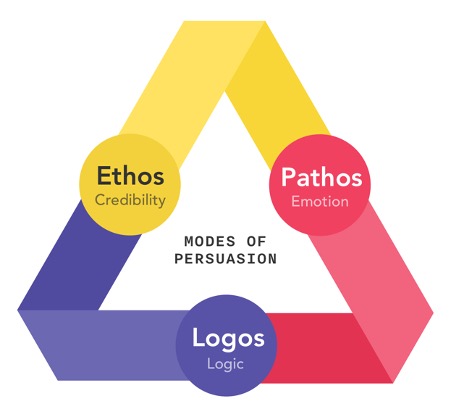}
\caption{The persuasion triangle, illustrating how logos, pathos, and ethos overlap and jointly contribute to persuasive message.}
\label{fig:persuasion_triangle}
\end{figure}

Persuasion Mode is the study of persuasive strategies categorized as Ethos, Logos, and Pathos, which are typically employed in persuasive tasks. First developed by Aristotle, Ethos refers to credibility, Logos represents reasoning, and Pathos pertains to emotion. As illustrated in Figure \ref{fig:persuasion_triangle}, these three modes overlap within a triangle, indicating that a persuasive text can incorporate one, two, or all three of them. For example, the statement, \textit{"Thousands of homeless children will sleep in the cold tonight unless we act now,"} has a strong emotional appeal that may compel a specific group to take immediate action. Conversely, another group might feel sympathy but prefer to suggest long-term policy solutions. This demonstrates that different persuasion modes resonate with people's various logical, emotional, and ethical reasoning.

\subsection{Computational Persuasion Mode Mining}

In the field of computational mining, the goal is to automatically identify persuasion structures and relationships. Argumentation mining and fallacy detection both require an understanding of persuasion modes. Research by \cite{higgins2012ethos, carlile2018give} examined Aristotelian persuasion strategies (i.e., Ethos, Logos, and Pathos), focusing primarily on reports and student essays in textual formats. However, they overlooked the potential of utilizing other modalities, such as images, which could enhance persuasiveness \citep{liu2022imagearg}. Although the persuasive mode has been studied in the audio-text modality within the context of fallacy detection shared tasks \citep{mancini2025overview}, this approach yielded poorer results compared to the purely textual modality. The authors highlighted several challenges, including noise that impeded the effective use of audio for this task. Consequently, we evaluate the computational paradigm of persuasion modes using the \texttt{ImageArg} dataset \citep{liu2022imagearg}, which incorporates images as an additional modality alongside text.



\subsection{Vision Language Models}

Vision-language models initially followed a similar approach to that of language models by using token masking during training \citep{bordes2024introduction}. The key difference was the incorporation of image processing through encoders, which extract features from images, while decoders primarily handle textual inputs. An important development in this area was the introduction of CLIP \citep{radford2021learning}, which proposed that both image and text modalities could share the same representational space, allowing them to enhance each other. Subsequent research aligned continuous image data with discrete textual data and introduced a new training strategy called Supervised Fine-tuning (SFT), where models are fine-tuned based on user instructions. As the field has evolved, there has been a concerted effort to optimize the loss function to effectively balance the relevant and irrelevant information during training, ultimately resulting in better generative models. These advancements have led to improved shape bias and alignment with human judgment \citep{jaini2024intriguing}. Consequently, VLMs have been extensively explored in simple visual question-answering tasks \citep{antol2015vqa}. However, there is limited literature on persuasion from a multi-modal perspective. A study by \citep{zhou2025joint} integrated pragma-dialectical steps with paralinguistic cues from audio elements into the in-context learning of the Qwen-3 reasoning model \citep{yang2025qwen3}. Nevertheless, they identified significant limitations in the models' ability to effectively process long prompts, which could affect their performance. To address this issue, we will use the coding manual provided by the \texttt{ImageArg} authors as our prompt. More details can be found in Section \ref{sec_label:experiment}.


%% file: sections/2_literature_review.tex
\section{Literature Review: ImageArg}

\subsection{Corpus Creation}

\begin{figure}[H]
\centering
\includegraphics[width=0.3\textwidth]{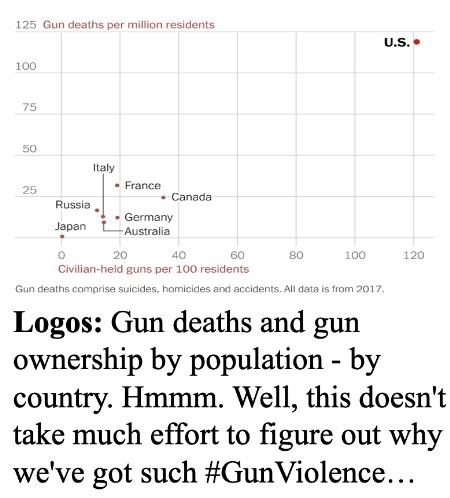}
\caption{Example of logos in the \texttt{ImageArg} dataset. The image presents statistical evidence on gun ownership and gun deaths across countries, while the accompanying tweet text interprets the observed relationship to support an argument about gun violence.}
\label{fig:eg_logos}
\end{figure}

\begin{figure}[H]
\centering
\includegraphics[width=0.3\textwidth]{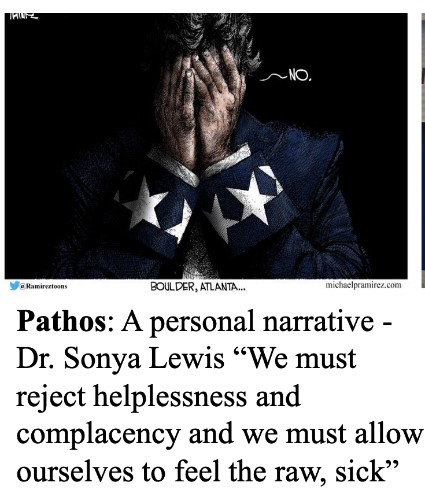}
\caption{Example of ethos in the \texttt{ImageArg} dataset. The image depicts Abraham Lincoln in an American-themed outfit, with his hand covering his face in grief. The accompanying tweet conveys a personal narrative from an independent individual asking for strength.}
\label{fig:eg_ethos}
\end{figure}

\begin{figure}[H]
\centering
\includegraphics[width=0.3\textwidth]{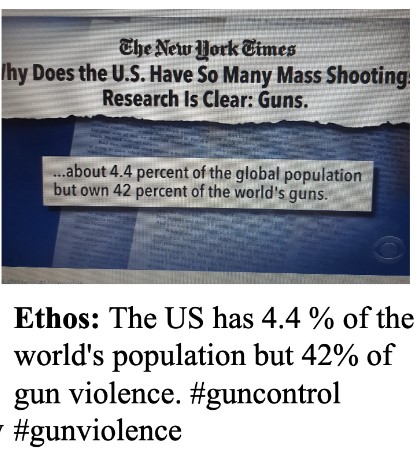}
\caption{Example of pathos in the \texttt{ImageArg} dataset. The image features a paper-cut headline from The New York Times alongside a paper-cut excerpt from the newspaper that provides facts supporting the headline. The accompanying tweets reiterate those facts.}
\label{fig:eg_pathos}
\end{figure}

The authors of the \texttt{ImageArg} dataset, as detailed in \cite{liu2022imagearg}, used the Twitter API\footnote{\url{https://developer.twitter.com/en/docs/twitter-api}} to gather multi-modal data related to gun control by employing specific keywords associated with the topic\footnote{The keywords were sourced from \cite{guo2020inflating}.}. They only retained tweets that received a confidence score greater than 0.9 from the ArgumentText Classify API\footnote{\url{https://api.argumentsearch.com}} for the purpose of annotation. Next, they set up a pipeline for annotating these instances, which included tasks ranging from basic stance detection to detailed persuasion mode detection. A notable aspect of their approach was the introduction of a metric called the \textit{Persuasive Score Improvement}. Annotators were required to first assign a score based solely on the text of the tweets and then provide a second score after reviewing the accompanying images. Only those scores that surpassed a predetermined threshold were considered valid for each detected persuasion mode. Ultimately, out of an initial total of 1,003 instances, only 259 were found suitable for the persuasion mode detection task. Table \ref{tab:image_arg_iaa} presents the inter-annotator agreement (IAA) scores \citep{krippendorff2011computing} for the persuasion mode, which ranged from 50\% to 58\%, indicating a moderate level of agreement among annotators. This variability highlights the challenges involved in the task, which are often influenced by the personal biases of the annotators. To enhance the effectiveness of the annotation process, the authors provided comprehensive instructions to the annotators. Figure \ref{fig:eg_logos}, \ref{fig:eg_ethos}, and \ref{fig:eg_pathos} each showcase examples of the different persuasion modes.

\input{tables/iaa_table}

\subsection{Multi-modal Benchmarking}

For benchmarking, \cite{liu2022imagearg} utilized the image encoder ResNet50 and the text encoder BERT to fine-tune linear classifiers. They referred to ResNet50 as I-M (Image Modality; \cite{he2016deep}) and BERT as T-M (Textual Modality; \cite{devlin2019bert}), with M-M representing multi-modality, where ResNet50 acts as the encoder and BERT as the decoder. They established a baseline with the random baseline, referred to as bm. The authors reported that bm achieved a higher F1 score in Logos and Pathos detection, while T-M performed competitively in Ethos detection. In terms of precision, T-M and I-M attained better scores in Pathos and Ethos detection, respectively. However, M-M showed poor performance across all tasks and evaluation metrics, suggesting that a more complex and better-trained M-M could lead to improved results. Subsequent works on this dataset focused exclusively on the stance detection task, raising the pressing question of:

(\textbf{Research Question}): \textit{How are current Vision-Language Models (VLMs) are evolving to address this complex challenge.}




%% file: tables/iaa_table.tex
\begin{table}[H]
    \centering
    \resizebox{0.3\columnwidth}{!}{%

     \begin{tabular}{ l c  c } 
     \toprule
      \textbf{Task} & \textbf{Alpha} & \textbf{Count}  \\
     \midrule
     \textbf{Logos} & 55.3 & 259 \\
     
     \textbf{Pathos} & 51.0 & 259 \\
     
     \textbf{Ethos} & 57.8 & 259 \\  

     \bottomrule
     
    \end{tabular}}
    \caption{Alpha illustrating the inter-annotator agreement (IAA) score of \texttt{ImageArg} on the Persuasion Mode task. }
    \label{tab:image_arg_iaa}
\end{table}

%% file: sections/3_experiment.tex
\section{Experiment, Results \& Discussion}



\subsection{Methodology}
\label{sec_label:experiment}

To answer this research question, we considered \textit{Qwen2-VL-7B-Instruct} and \textit{Qwen3-VL-8B-Instruct} of the same Qwen family. We set the models that achieved the best F1-score on respective persuasion modes on the ImageArg dataset as baseline, i.e., (bm for Logos and Pathos, and T-M for Ethos). We only evaluated on the test set on a zero-shot setting and adopted the same evaluation metric of this paper, i.e., Precision, Recall, and F1-score, and kept the same image size reported in the paper for fair evaluation. For prompt, we utilized the code manual and instructions used to build this dataset during annotation and set \texttt{generation\_prompt} as \texttt{TRUE}. We left the remaining generation parameters of these models unchanged and ran on a single NVIDIA A100 Tensor Core GPU.




\subsection{Results}

\input{tables/results}

Table \ref{tab:persuasion_results} presents the results of this study. Overall, the Qwen models achieved the highest F1-scores across all persuasion modes. Specifically, the Qwen models demonstrated significantly improved precision in detecting Logos, indicating a better understanding of logical reasoning as it relates to interpreting statistical evidence in images. Additionally, both models showed competitive results in terms of precision and recall, with Qwen2 performing slightly better in comprehending persuasive texts and images. Interestingly, Ethos is the only task where both models were very close to the baseline, highlighting the challenge of overcoming emotional bias, which also influenced the annotation process. 

\subsection{Discussion}

To compare the two generations of the Qwen family, Qwen3 exhibited superior ethical and credibility comprehension compared to Qwen2. This suggests that Qwen3 has enhanced image processing capabilities and better utilization of numerical image data, while Qwen2 slightly excels in understanding hidden emotional appeals. Looking ahead, it would be interesting to investigate how much image modality influenced these performances, which closely aligns with this paper's other task, \textit{Persuasive Score Improvement}.

%% file: tables/results.tex
\begin{table}[H]
\centering
\small
\begin{tabular}{l l cccc}
\toprule
\textbf{Persu. Mode} & \textbf{Model} & Prec ↑ & Rec ↑ & F1 ↑\\
\midrule

\multirow{2}{*}{Logos}
& BASE (\textit{bm}) & 0.405 & \textbf{1.000} & 0.575\\
& Qwen2 & 0.596 & 0.885 & \textbf{0.709}\\
& Qwen3 & \textbf{0.633} & 0.813 & \textbf{0.709}\\
\midrule

\multirow{2}{*}{Pathos}
& BASE (\textit{bm}) & 0.554 & \textbf{1.000} & 0.712 \\
& Qwen2 & \textbf{0.568} & 0.969 & \textbf{0.714} \\
& Qwen3 & 0.538 & \textbf{1.000} & 0.696 \\
\midrule

\multirow{2}{*}{Ethos}
& T-M (\textit{bm}) & 0.168 & 0.817 & 0.272 \\
& Qwen2 & 0.195 & 0.550 & 0.277 \\
& Qwen3 & \textbf{0.236} & \textbf{0.875} & \textbf{0.365} \\

\bottomrule
\end{tabular}
\caption{Performance of persuasion mode classification across Logos, Pathos, and Ethos using precision (Prec), recall (Rec), and F1-score metrics. The Qwen models achieved the strongest overall performance, attaining the highest F1-scores across persuasion modes. Bold values indicate the best result for each metric within a persuasion mode.}
\label{tab:persuasion_results}
\end{table}

%% file: sections/5_conclusion.tex
\section{Conclusion}

In this research, we evaluate the effectiveness of Vision-Language Models (VLMs) in the complex task of detecting persuasion modes. Persuasion modes—Logos, Pathos, and Ethos—overlap within a persuasion triangle, complicating the influence of inherent personal biases. For our evaluation, we selected the \texttt{ImageArg} dataset, where the authors reported moderate annotator agreement due to the challenges associated with the task. To assess model performance, we focused on two generation models from the Qwen family to understand how these models are evolving to tackle this complex challenge. Overall, we found that Qwen models achieved the highest F1 scores when compared to benchmark models. Specifically, Qwen3 demonstrated superior performance in identifying the Logos and Pathos modes, while Qwen2 performed competitively in the more difficult Ethos detection task. Future research could explore the effectiveness of these VLMs in detecting more nuanced persuasion modes, such as \textit{Ad Hominem} and \textit{Appeal to Authority}, as well as their application to related tasks such as Fallacy Detection.



%% file: sections/6_limitations.tex
\section{Limitations}
\subsection{Alternative VLM Family}
Currently, numerous VLM model families are available to address this research question. However, we utilize Qwen because of its popularity and comprehensive open-source report. Since different VLMs are trained with distinct objectives, their performance may vary if another VLM were chosen.

\subsection{Evaluation Setup}
In our study, we evaluate only in a zero-shot setting. Since the community representing these language families does not release their training data, we assume that the training instances for these tasks were included in the models' training. However, this may not necessarily be the case. If it is not, an n-shot setup could be employed to compare the results against the same baseline, which was trained with significantly more samples.

%% file: termPaper.bbl
\begin{thebibliography}{}

\bibitem[Antol et~al., 2015]{antol2015vqa}
Antol, S., Agrawal, A., Lu, J., Mitchell, M., Batra, D., Zitnick, C.~L., and Parikh, D. (2015).
\newblock Vqa: Visual question answering.
\newblock In {\em Proceedings of the IEEE international conference on computer vision}, pages 2425--2433.

\bibitem[Bordes et~al., 2024]{bordes2024introduction}
Bordes, F., Pang, R.~Y., Ajay, A., Li, A.~C., Bardes, A., Petryk, S., Ma{\~n}as, O., Lin, Z., Mahmoud, A., Jayaraman, B., et~al. (2024).
\newblock An introduction to vision-language modeling.
\newblock {\em arXiv preprint arXiv:2405.17247}.

\bibitem[Carlile et~al., 2018]{carlile2018give}
Carlile, W., Gurrapadi, N., Ke, Z., and Ng, V. (2018).
\newblock Give me more feedback: Annotating argument persuasiveness and related attributes in student essays.
\newblock In {\em Proceedings of the 56th Annual Meeting of the Association for Computational Linguistics (Volume 1: Long Papers)}, pages 621--631.

\bibitem[Devlin et~al., 2019]{devlin2019bert}
Devlin, J., Chang, M.-W., Lee, K., and Toutanova, K. (2019).
\newblock Bert: Pre-training of deep bidirectional transformers for language understanding.
\newblock In {\em Proceedings of the 2019 conference of the North American chapter of the association for computational linguistics: human language technologies, volume 1 (long and short papers)}, pages 4171--4186.

\bibitem[Guo et~al., 2020]{guo2020inflating}
Guo, M., Hwa, R., Lin, Y.-R., and Chung, W.-T. (2020).
\newblock Inflating topic relevance with ideology: A case study of political ideology bias in social topic detection models.
\newblock In {\em Proceedings of the 28th International Conference on Computational Linguistics}, pages 4873--4885.

\bibitem[He et~al., 2016]{he2016deep}
He, K., Zhang, X., Ren, S., and Sun, J. (2016).
\newblock Deep residual learning for image recognition.
\newblock In {\em Proceedings of the IEEE conference on computer vision and pattern recognition}, pages 770--778.

\bibitem[Higgins and Walker, 2012]{higgins2012ethos}
Higgins, C. and Walker, R. (2012).
\newblock Ethos, logos, pathos: Strategies of persuasion in social/environmental reports.
\newblock In {\em Accounting forum}, volume~36, pages 194--208. Elsevier.

\bibitem[Jaini et~al., 2024]{jaini2024intriguing}
Jaini, P., Clark, K., and Geirhos, R. (2024).
\newblock Intriguing properties of generative classifiers.
\newblock In {\em International Conference on Learning Representations}, volume 2024, pages 13898--13923.

\bibitem[Krippendorff, 2011]{krippendorff2011computing}
Krippendorff, K. (2011).
\newblock Computing krippendorff's alpha-reliability.

\bibitem[Liu et~al., 2022]{liu2022imagearg}
Liu, Z., Guo, M., Dai, Y., and Litman, D. (2022).
\newblock Imagearg: A multi-modal tweet dataset for image persuasiveness mining.
\newblock In {\em Proceedings of the 9th Workshop on Argument Mining}, pages 1--18.

\bibitem[Mancini et~al., 2025]{mancini2025overview}
Mancini, E., Ruggeri, F., Villata, S., and Torroni, P. (2025).
\newblock Overview of mm-argfallacy2025 on multimodal argumentative fallacy detection and classification in political debates.
\newblock In {\em Proceedings of the 12th Argument mining Workshop}, pages 358--368.

\bibitem[Radford et~al., 2021]{radford2021learning}
Radford, A., Kim, J.~W., Hallacy, C., Ramesh, A., Goh, G., Agarwal, S., Sastry, G., Askell, A., Mishkin, P., Clark, J., et~al. (2021).
\newblock Learning transferable visual models from natural language supervision.
\newblock In {\em International conference on machine learning}, pages 8748--8763. PmLR.

\bibitem[Yang et~al., 2025]{yang2025qwen3}
Yang, A., Li, A., Yang, B., Zhang, B., Hui, B., Zheng, B., Yu, B., Gao, C., Huang, C., Lv, C., et~al. (2025).
\newblock Qwen3 technical report.
\newblock {\em arXiv preprint arXiv:2505.09388}.

\bibitem[Zhou et~al., 2025]{zhou2025joint}
Zhou, H., Westerdijk, H., and Islam, K.~I. (2025).
\newblock Joint effects of argumentation theory, audio modality and data enrichment on llm-based fallacy classification.
\newblock {\em arXiv preprint arXiv:2509.11127}.

\end{thebibliography}
